\documentclass[11pt,a4paper]{article}
\usepackage[hyperref]{acl2017}
\usepackage{times}
\usepackage{latexsym}

\usepackage{url}

\usepackage{amsmath,amsfonts,amsthm,bm} 
\newcommand{\bs}{\boldsymbol}
\usepackage{graphicx}
\usepackage{lipsum}
\allowdisplaybreaks
\usepackage{cleveref}
\usepackage{subfigure}
\usepackage{caption}

\aclfinalcopy % Uncomment this line for the final submission
\title{Incorporating Syntactic Uncertainty in Neural Machine Translation with Forest-to-Sequence Model}

\author{Poorya Zaremoodi \hspace{1.5cm}  Gholamreza Haffari\\
  Faculty of Information Technology, Monash University \\
  {\tt first.last@monash.edu} \\}

\date{}

\begin{document}
\maketitle

\begin{abstract}
Incorporating syntactic information in Neural Machine Translation models is a method to compensate their requirement for a large amount of parallel training text, specially for low-resource language pairs. Previous works on using syntactic information provided by (inevitably error-prone) parsers has been promising. In this paper, we propose a forest-to-sequence Attentional Neural Machine Translation model to make use of exponentially many parse trees of the source sentence to compensate for the parser errors. Our method represents the collection of parse trees as a packed forest, and learns a neural attentional transduction model from the forest to the target sentence. Experiments on English to German, Chinese and Persian translation show the superiority of our method over the tree-to-sequence and vanilla sequence-to-sequence neural translation models.
\end{abstract}

\section{Introduction}
Recently, a great deal of attention has been focused on Neural Machine Translation (NMT). The main NMT approach is based on the encoder-decoder architecture \cite{cho2014learning, sutskever2014sequence} in which an encoder  (e.g an RNN) reads the source sentences sequentially to produce a fixed-length vector representation. Then, a decoder generates a translation from the encoded vector. Attention mechanism is one of the most successful extensions to the simple encoder-decoder models. Attentional models jointly learn to softly align to the source sentence while they attempt to generate words in the target language. 

In a natural language, the relationship between words is considered to be in a latent nested hierarchy rather than a sequence order. Therefore, it is expected that modelling these hierarchies (e.g. syntactic tree) will improve the performance of the NMT model, specially for linguistically distant language pairs like English $\rightarrow$ Farsi. Incorporating syntactic constituents of the source language has been studied in Statistical Machine Translation and it has improved translation accuracy. Recently, \citep{eriguchi2016tree} proposed a method to incorporate the hierarchical syntactic information of the source sentence. In addition to the embedding of words which is computed using the basic attentional method, they compute the embedding of phrases in a bottom-up fashion, directed by a parse tree of the source sentence. Since generating gold-standard parse trees is expensive, they used trees generated by off-the-shelf parsers. However, using 1-best parse tree can lead to translation mistakes due to parsing errors.

In this paper, we address the aforementioned issue by using a compact forest instead of a single 1-best tree. In this approach, the embeddings of phrases are calculated using a bottom-up fashion using dynamic programming based on the exponentially many trees encoded in the forest. Thus, in the encoding stage of this approach, different ways of constructing a phrase can be taken into consideration along with the probability of rules in the corresponding trees. Evaluation of this approach on English to Chinese, Persian and German translation showed that forests has better performance compared to 1-best tree and sequential ANMT models.

\section{Neural Machine Translation}

\subsection{Attentional Sequence-to-Sequence model}\label{sec:vanAtt}
\begin{figure}
\centering
\includegraphics[scale=0.5]{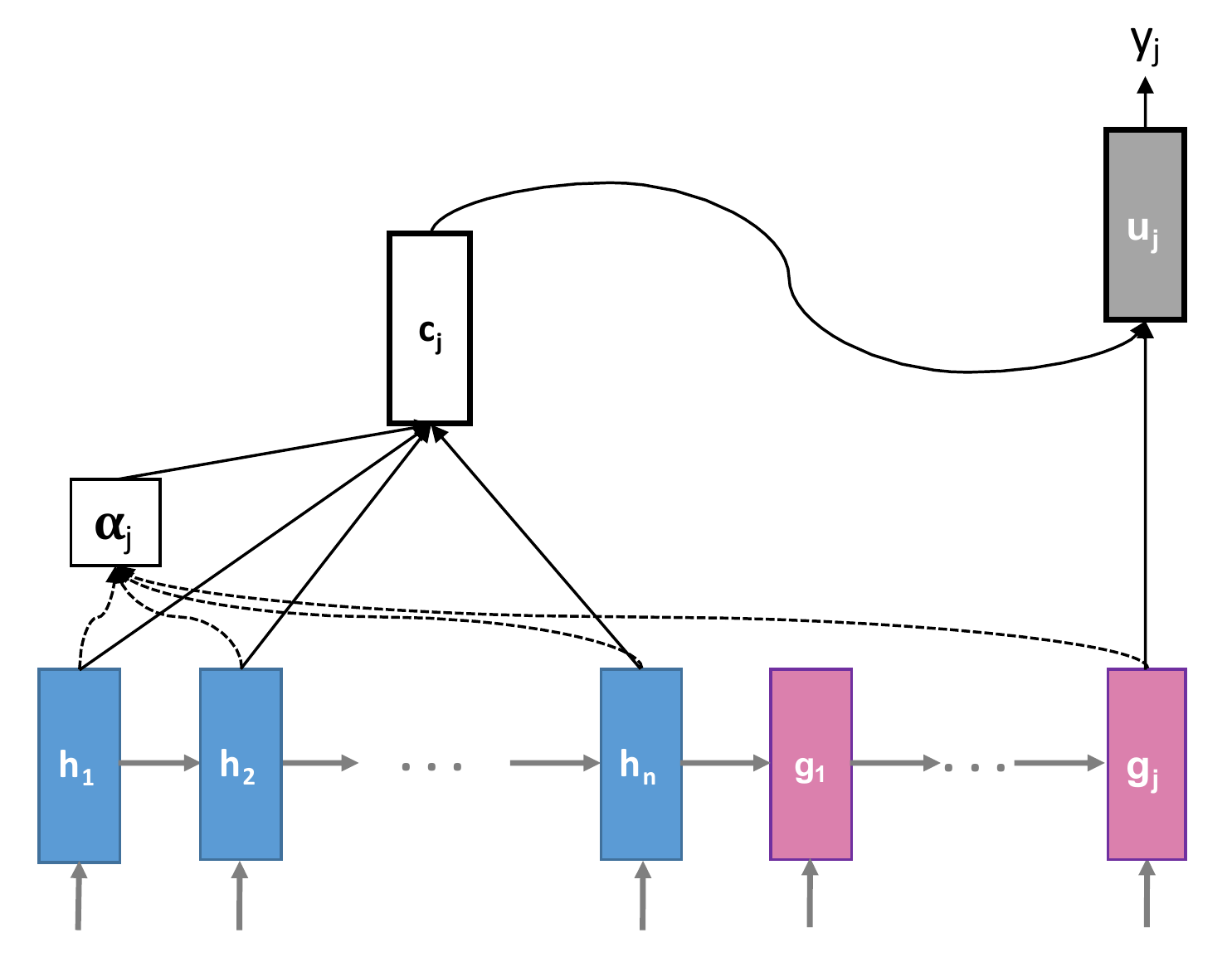}
\caption{Attentional Encoder-Decoder model.}\label{fig:vanillaAtt}
\end{figure}

The architecture of this approach is depicted in Figure \ref{fig:vanillaAtt}  .
\subsubsection{Encoder}\label{sec:vanEnc}
The source sentence is encoded using an RNN. The hidden states of these RNNs are computed as follows:
\begin{gather*}
\bs{h}_i = f_{enc}(\bs{h}_{i-1},E_x[\bs{x}_{i-1}])
\end{gather*}
where $E_x$ is embedding lookup table of the source word $x$. We use Long Short-Term Memory (LSTM) unit \cite{hochreiter1997long} as $f_{enc}$.
Context-dependent embedding of each word is the corresponding hidden state of the encoder ($\bs{e}_i=\bs{h}_i$).

\subsubsection{Decoder}\label{sec:vanDec}
The decoder generates output using a deep output RNN \cite{pascanu2013construct} which is an RNN followed by an MLP (1-layer in the basic version) to generate the next output word in each time-step. The decoder also consists of an attention mechanism to perform soft alignment to the computed embeddings of the source words.

For the generation of each word $j$, the attention mechanism generates a context vector $\bs{c}_j$ from the source sentence and with respect to the current state of the decoder. Then, the decoder produce the $j$-th word with the following equations:
\begin{gather}
\bs{g}_j = \tanh (\bs{W}^{gh}\bs{g}_{j-1}+\bs{W}^{gi} \bs{E}_y[\bs{y}_{j-1}]  \notag \\ +\bs{W}^{ga}\bs{c}_j +\bs{W}^{gu}\bs{u}_{j-1}), \label{eq:vanG}  \\
\bs{u}_j = \tanh (\bs{W}^{uc}\bs{c}_{j}+\bs{W}^{ui} \bs{E}_y[\bs{y}_{j-1}]+\bs{g_j}), \label{eq:vanU} \\ 
p(y_j| \bs{y}_{< j},\bs{x}) =  \mbox{softmax} (\bs{W}^{ou}\bs{u_{j}}+\bs{b}^{o}), \label{eq:vanO}
\end{gather}
where $\bs{E}_y[y_j]$ is the embedding of word $y_j$ looked-up from target language embedding table $\bs{E}_y$, and $\bs{W}$ matrices and $\bs{b}^o$ are parameters. The last term in the Equation \ref{eq:vanG} is the input-feeding term \cite{luong2015effective}.

A crucial part of the decoder is the attention mechanism which provide relevant information from the source sentence with respect to the current state of the decoder.
The context vector is calculated as follows:
\begin{align*}
a_{ij} &= \bs{v}^\top \tanh (\bs{W}^{ae}\bs{e}_i + \bs{W}^{ag}\bs{g}_{j-1}), \\
\bs{\alpha}_{j} &= \text{softmax} (\bs{a}_j), \\
\bs{c}_j &=\sum_i \alpha_{ji}\bs{e}_i,
\end{align*}
where $a_{ij}$ is the score determining how well the words around position $i$ in the source sentence match the state of the decoder before generating the $j$-th word.

Training of this model is done by minimising the following objective function:
\begin{equation*}
J(\theta)=\dfrac{1}{|D|}\sum_{(\bs{x},\bs{y})\in D} -\log p(\bs{y}|\bs{x})
\end{equation*}
where $D$ is the set of bilingual training sentence pairs.

\subsection{Attentional Tree-to-Sequence model}

\begin{figure}
\centering
\includegraphics[scale=0.5]{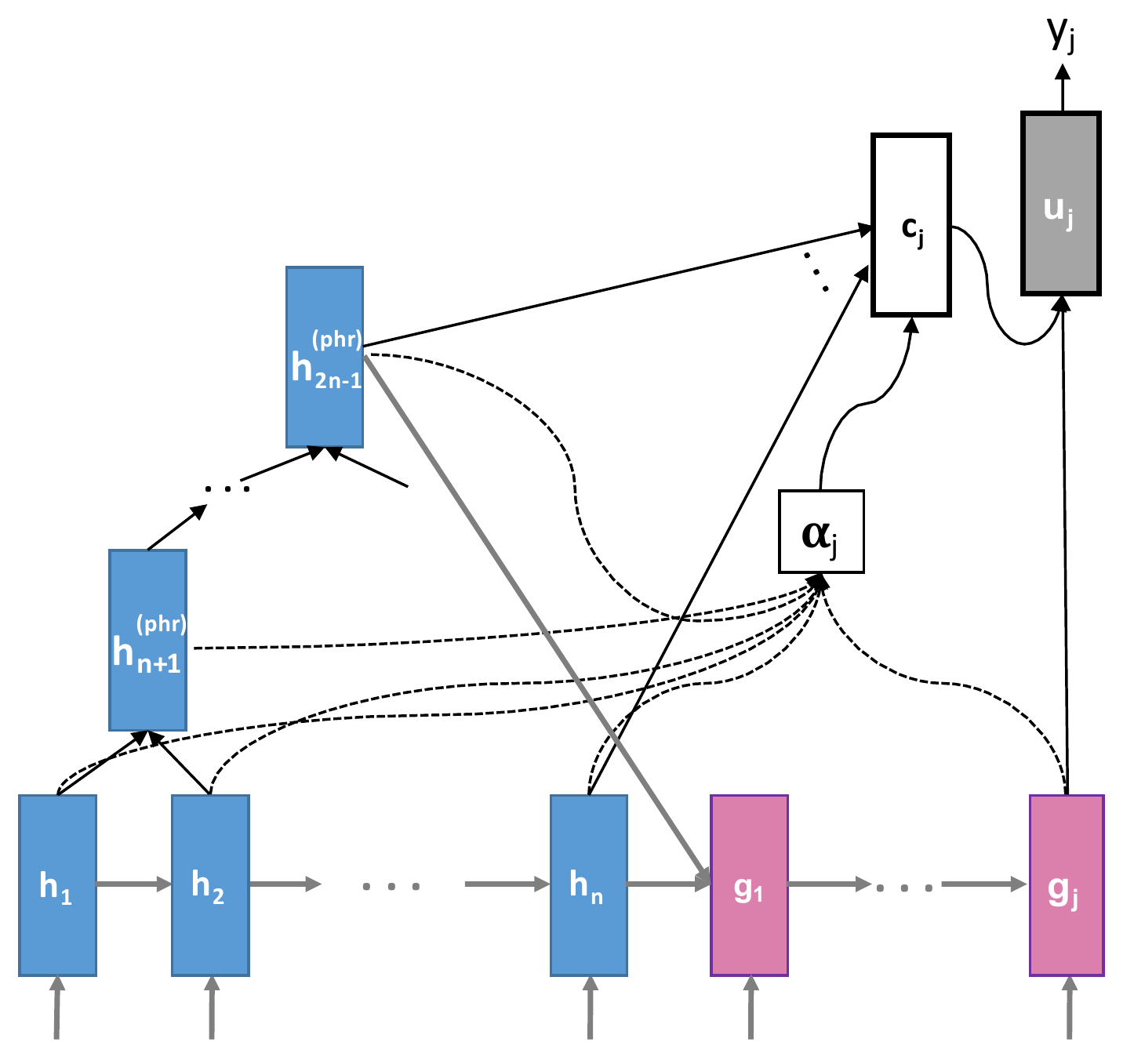}
\caption{Attentional Tree-to-Sequence model.}\label{fig:treeAtt}
\end{figure}

\citep{eriguchi2016tree} proposed a method to incorporate the hierarchical syntactic information of the source sentence. They compute embeddings of phrases in addition to the words, then attend to both words and phrases in the decoder. This process is depicted in Figure \ref{fig:treeAtt}. Phrase embeddings are computed in a bottom-up fashion directed by the parse tree of source sentence. Since generating gold-standard parse trees is expensive, they used binary constituency parse trees generated by a Head-driven Phrase Structure Grammar (HPSG) parser.

\subsubsection{Encoder}\label{sec:treeEnc}
The encoding process is done in two phases. In the first phase, a standard sequential encoder (Section \ref{sec:vanAtt}) is operated on the input sequence. Then a tree-based encoder is operated on the embeddings resulted by the sequential encoder\footnote{This model uses one-directional LSTM in the sequential encoder}. The embedding of node $k$, $\bs{h}_k^{(phr)}$, is calculated by combining the embeddings of its children using a non-linear function. Since, the tree is binary, the combining operation is as follow:
\begin{equation*}
\bs{h}_k^{(phr)} = f_{tree}(\bs{h}_k^l, \bs{h}_k^r).
\end{equation*}
where $f_{tree}$ is a non-linear function, $\bs{h}_k^l$ and $\bs{h}_k^r$ are hidden states of left and right children respectively. This method uses Tree-LSTM components \cite{tai2015improved} to calculate the LSTM unit of the parent node using the two children units as follow:
\begin{align*}
\bs{i} &= \sigma (\bs{U}_l^{(i)}\bs{h}^l + \bs{U}_r^{(i)}\bs{h}^r  + \bs{b}^{(i)})\\
\bs{f}^l &=\sigma (\bs{U}_l^{(f_l)}\bs{h}^l + \bs{U}_r^{(f_l)}\bs{h}^r  + \bs{b}^{(f_l)})\\
\bs{f}^r &=\sigma (\bs{U}_l^{(f_r)}\bs{h}^r + \bs{U}_r^{(f_r)}\bs{h}^r  + \bs{b}^{(f_r)})\\
\bs{o} &= \sigma (\bs{U}_l^{(o)}\bs{h}^l + \bs{U}_r^{(o)}\bs{h}^r  + \bs{b}^{(o)})\\
\tilde{\bs{c}} &= \tanh (\bs{U}_l^{(\tilde{c})}\bs{h}^l + \bs{U}_r^{(\tilde{c})}\bs{h}^r  + \bs{b}^{(\tilde{c})})\\
\bs{c}^{(phr)} &= \bs{i} \odot \tilde{\bs{c}} + \bs{f}^l \odot \bs{c}^l + \bs{f}^r \odot \bs{c}^r\\
\bs{h}^{(phr)} &= \bs{o} \odot \tanh (\bs{c}^{(phr)})
\end{align*}
where $\bs{i}$, $\bs{f}^l$, $\bs{f}^r$, $\bs{o}_j$,$\bs{\tilde{c}}_j$ are the input gate, left and right forget gates, output gate, and a state for updating memory cell, and $\bs{c}^r$ and $\bs{c}^l$ are memory cells of the right and left units. 
\subsubsection{Decoder}\label{sec:treeDec}
The initial state of the decoder is calculated by combining the final state of sequential and tree encoders as follow:
\begin{equation*}
\bs{g}_1 = g_{tree}(\bs{h}_n, \bs{h}_{root}^{(phr)}),
\end{equation*}
where $g_{tree}$ is a function similar to $f_{tree}$ with another set of parameters. The rest of the decoder is similar to the vanilla attentional decoder discussed in Section \ref{sec:vanDec}. The difference is that, in this model, the attention mechanism makes use of phrases as well as words. Thus, the context component is calculated as follows:
\begin{equation*}
\bs{c}_j=\sum_{i=1}^{n} \alpha_{ji}\bs{h}_i + \sum_{i=n+1}^{2n-1} \alpha_{ji}\bs{h}_i^{(phr)}
\end{equation*}

\section{Forest-to-Sequence Attentional Encoder-Decoder Model}
The tree-to-sequence model uses 1-best parse tree generated by a parser. Mistakes and uncertainty in parsing eventually affect the performance of the translation.  In this section we propose a novel ANMT model which utilises forest of trees. Forest-to-sequence model computes embeddings of phrases using a bottom-up fashion based on exponentially many trees instead of 1-best one. Calculating embeddings of phrases using each tree separately results in an exponential time complexity. To reduce computational complexity, we use forest data structure and dynamic programming for the computation of phrase embeddings. We use the packed forests generated by \cite{huang2008forest}\footnote{generated trees are not necessarily binary} which is an extension to the Charniak parser \cite{charniak2005coarse}. Tags are discarded, and just the structures of trees are taken into consideration.

\begin{figure}
\centering
\includegraphics[scale=0.63]{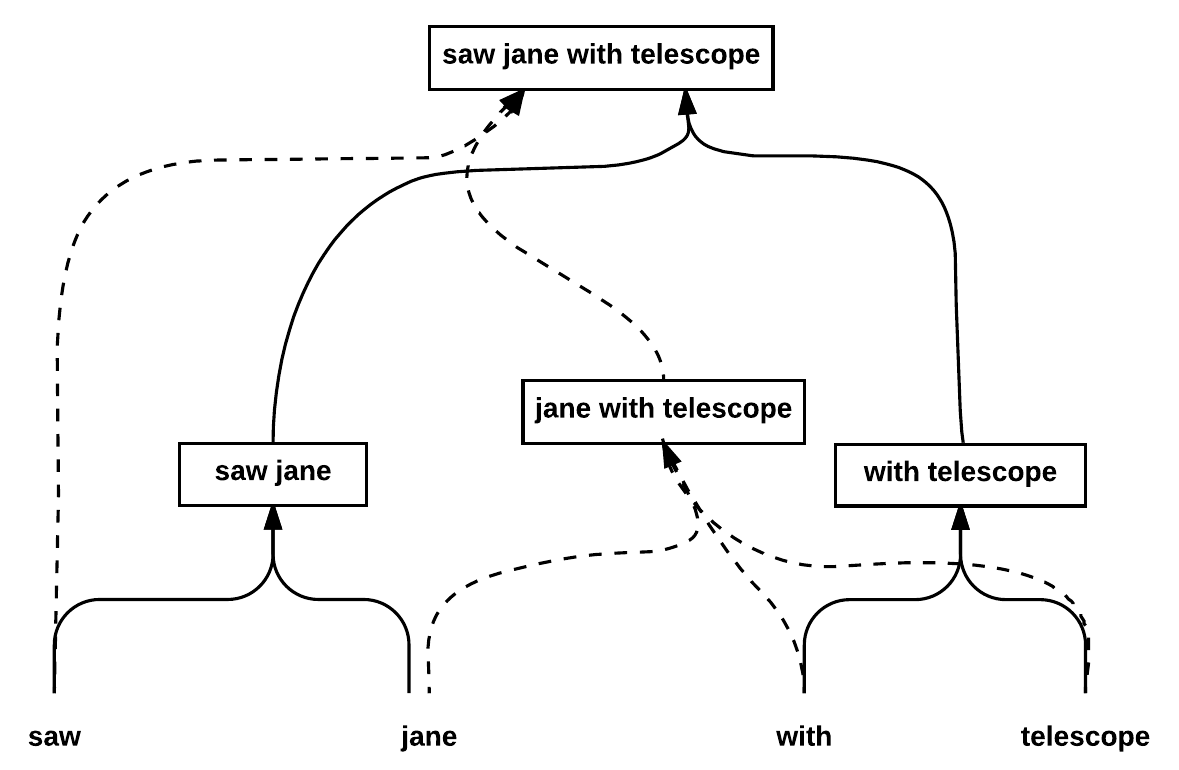}
\caption{An example of generating a phrase from two different parse trees}\label{fig:forestAtt}
\end{figure}

Figure \ref{fig:forestAtt} depicts two different parse trees for a phrase. As seen, a phrase can be constructed in more than one way. Thus, multiple embeddings can be computed for a phrase. In order to make computation of phrase embeddings more efficient, we combine all different embeddings of a phrase to get a unified embedding for each phrase.  In upper levels of the forest, we just use the unified embedding of phrases in lower levels.

In order to compute embeddings of words, a standard sequential encoder is operated on the input sequence to obtain embeddings of words. Then, we compute the embedding of each phrase in two phases. First, if one or more trees generate a phrase, for each tree we calculate the intermediate LSTM representation of the phrase using the tree-LSTM encoder. Then, we combine these intermediate representations using the proposed forest-LSTM encoder to obtain the final representation of the phrase. Our forest-LSTM combines phrase embeddings resulted of multiple trees to obtain a unified embedding for the phrase as follows:

{\small
\begin{align*}
\bs{\gamma}^l &= \tanh (\bs{U}^{(\gamma)}\sum_{l'=1}^{N}\bs{1}_{l\neq l'}\bs{h}^{l'} +\bs{W}^{(\gamma)}\bs{h}^l + \bs{v}^{(\gamma)} p^l +\bs{b}^{(\gamma)})  \\
\bs{i} &= \sigma (\bs{U}^{(i)}\sum_{l=1}^{N}[\bs{h}^l;\bs{\gamma}^l] + \bs{b}^{(i)})\\
\bs{f}^l &=\sigma (\bs{U}^{(f)}\sum_{l'=1}^{N}\bs{1}_{l\neq l'}[\bs{h}^{l'};\bs{\gamma}^{l'}]+ \bs{W}^{(f)}[\bs{h}^l;\bs{\gamma}^l] + \bs{b}^{(l)}) \\
\bs{o} &= \sigma (\bs{U}^{(o)}\sum_{l=1}^{N}[\bs{h}^l;\bs{\gamma}^l]  + \bs{b}^{(o)})\\
\tilde{\bs{c}} &= \tanh (\bs{U}^{(\tilde{c})}\sum_{l=1}^{N}[\bs{h}^l ;\bs{\gamma}^l] + \bs{b}^{(\tilde{c})})\\
\bs{c}^{(phr)} &= \bs{i} \odot \tilde{\bs{c}} + \sum_{l=1}^{N}\bs{f}^l \odot \bs{c}^l\\
\bs{h}^{(phr)} &=\bs{o} \odot \tanh (\bs{c}^{(phr)})
\end{align*}}
where $N$ is the number of children which in fact shows different ways to generate the current phrase, and $p^l$ is the probability of $l$-th child which is the probability of the rule for generating the phrase in a corresponding tree. We use an additional layer to compute embedding vector for probability of each child ($\bs{\gamma}^{(.)}$) because it helps to impose probabilities into gates with respect to the hidden states and probability of all children. Also, in computation of probability embedding and forget gate of a child, we used different weight matrices for the current child ($\bs{U}$) and others ($\bs{W}$).

For the decoding part, we used vanilla attentional decoder discussed in Section \ref{sec:vanDec} with this change that the attention mechanism attends on both words and phrases. In each decoding step, context component is calculated as follows:
\begin{equation*}
\bs{c}_j=\sum_{i=1}^{n} \alpha_{ji}\bs{h}_i + \sum_{i=n+1}^{N_p} \alpha_{ji}\bs{h}_i^{(phr)}
\end{equation*}
where $N_p$ is the number of different phrases in the forest. In contrast with tree-to-sequence model, in our method the number of embedded phrases are depend on parser and its ambiguities during parsing. Also, the probability of each phrase is taken into consideration.
\color{black}
\section{Experiments}

\begin{table*}[t]
\begin{minipage}{\textwidth}
\centering
\begin{tabular}{|c|c|c|c|c|c|c|c|}

\multicolumn{2}{c}{ }& \multicolumn{2}{c}{En $\rightarrow$ De}  & \multicolumn{2}{c}{En $\rightarrow$ Ch} & \multicolumn{2}{c}{En $\rightarrow$ Fa}  \\
\cline{2-8}
\multicolumn{1}{c|}{Method}  & H &  Perplexity & BLEU & Perplexity & BLEU & Perplexity & BLEU  \\
\hline
\hline
Proposed & 512  & 29.25 & 13.43 & 5.49 & 28.39 & 16.66 & 12.38   \\
Forest-to-sequence & 256 & 30.83 & 13.54 & 6.16 & 27.08 & 17.62 & 11.91   \\
\hline
Tree-to-sequence & 512 & 31.86 & 13.05 & 5.71 & 28 & 16.28 & 11.71  \\
 \cite{eriguchi2016tree} & 256 & 30.13 & 13 & 6.17 & 26.85 & 17.94 & 11.32   \\
\hline
Vanilla ANMT  & 512 & 32.61 & 12.21 & 6.12	 & 26.77 & 18.4 & 10.93   \\
 \cite{luong2015effective}  & 256 & 33.07 & 11.98 & 6.48 & 25.43 & 19.21 & 10.17   \\

\hline
\end{tabular}
\caption{Comparison of the methods together with different hidden dimension size (H) for all datasets.}\label{tab:res}
\end{minipage}
\end{table*}

\subsection{Datasets}
We make use of three different language pairs, English (En) $\rightarrow$ Chinese (Ch)/ Persian (Fa)/ German (De). For English$\rightarrow$Chinese, we use the BTEC corpus which has 44,016 sentence pairs for training where 'devset1\_2' and 'devset\_3' are used for development and test purpose, respectively. For En $\rightarrow$ Fa we used TEP dataset \cite{tiedemann2009news} that is extracted by aligning English and Persian movie subtitles. It has about 341K sentence pairs. We split the data in 337K, 2K, 2K parts for training, development and test purpose respectively. For En $\rightarrow$ De, we use the first 100K sentences of Europarl v7 part of WMT’14 \footnote{http://www.statmt.org/wmt14/translation-task.html} training data, and newstest2013 and newstest2014 are used as development and test sets, respectively.
 
In the preprocessing stage, datasets are lower-cased and tokenised using the Moses toolkit \cite{moses}. Sentences longer than 50 words are removed, and words with frequency less than 5 are converted to a special $<unk>$ symbol. Compact forests and tress of English side sentences are obtained from \cite{huang2008forest} parser. In all experiments, we used development data just for model selection, and test data for evaluation purposes.

\textbf{Training details}
We use \cite{cohn2016incorporating} implementation of the ANMT model, and implement  the tree-to-sequence and our model using the DyNet toolkit \cite{dynet} on top of it. Models are trained end-to-end using Stochastic Gradient Descent (SGD) where the mini-batch size was is to 128. Each model is trained up to 20 epochs on a GPU\footnote{Nvidia Tesla K20m}. Validation-based early stopping (on development sets) is applied to regularise the model and prevent overfitting. The translation is decoded greedy, and BLEU score is computed using 'multi-bleu.perl' script of Moses \cite{moses} toolkit. 

\subsection{Experimental Results}
The perplexity and BLEU score of models for all translation tasks are presented in Table \ref{tab:res}. In all cases, our work achieved higher BLEU score (up to ~0.7 score) compared to the two other models for all of the datasets. Therefore, it can be inferred that using grammar forests can compensate the errors and ambiguities of parsers.

\begin{figure*}[!t]
\centering
\subfigure[]{\label{fig:bg}\includegraphics[width=0.3\textwidth]{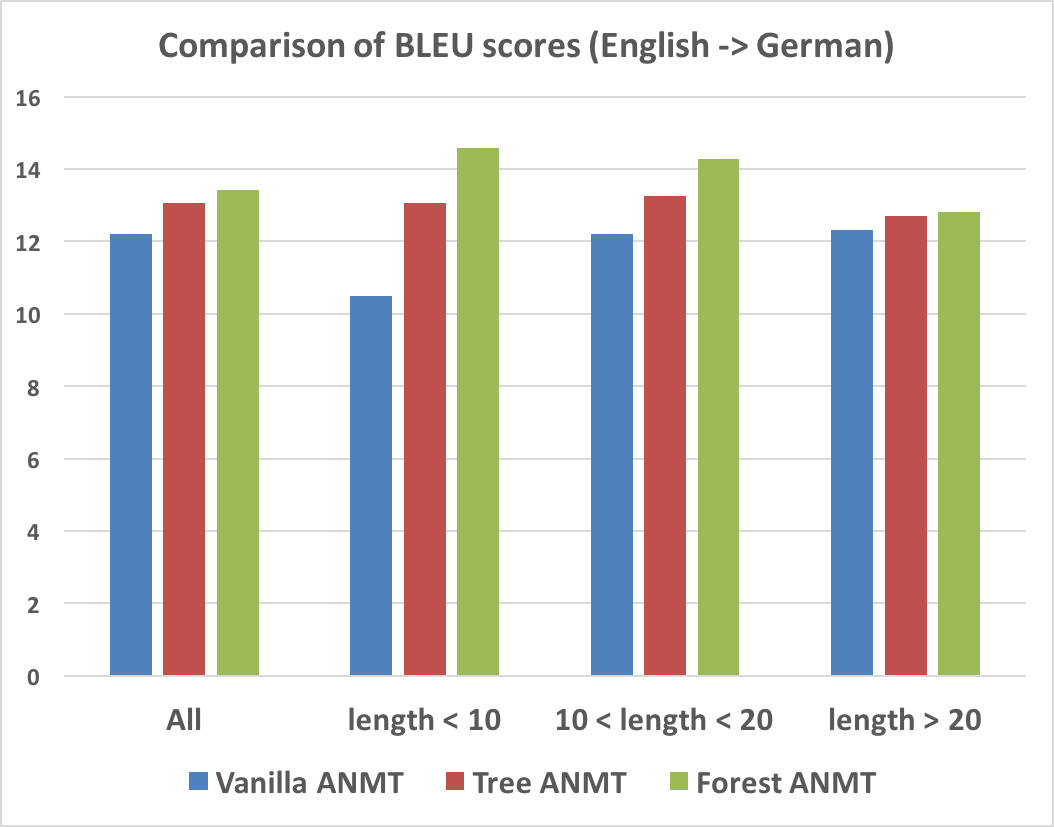}}
\hfill 
\subfigure[]{\label{fig:bc}\includegraphics[width=0.3\textwidth]{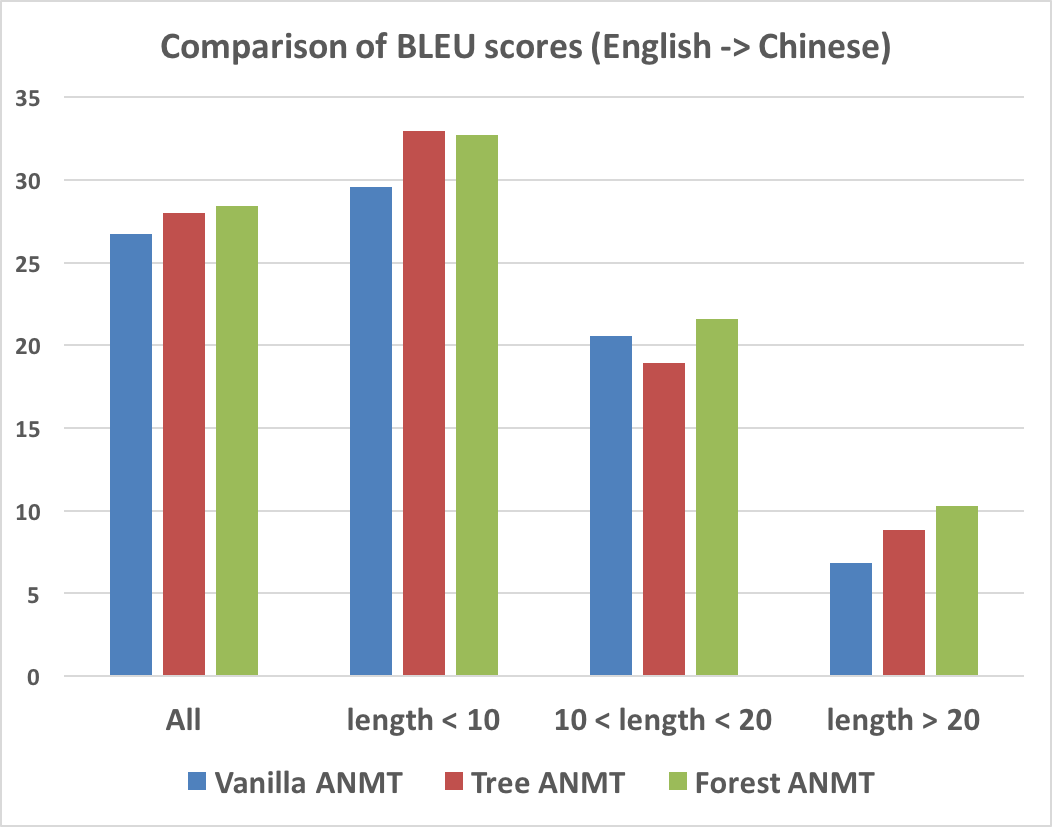}} 
\hfill
\subfigure[]{\label{fig:bf}\includegraphics[width=0.3\textwidth]{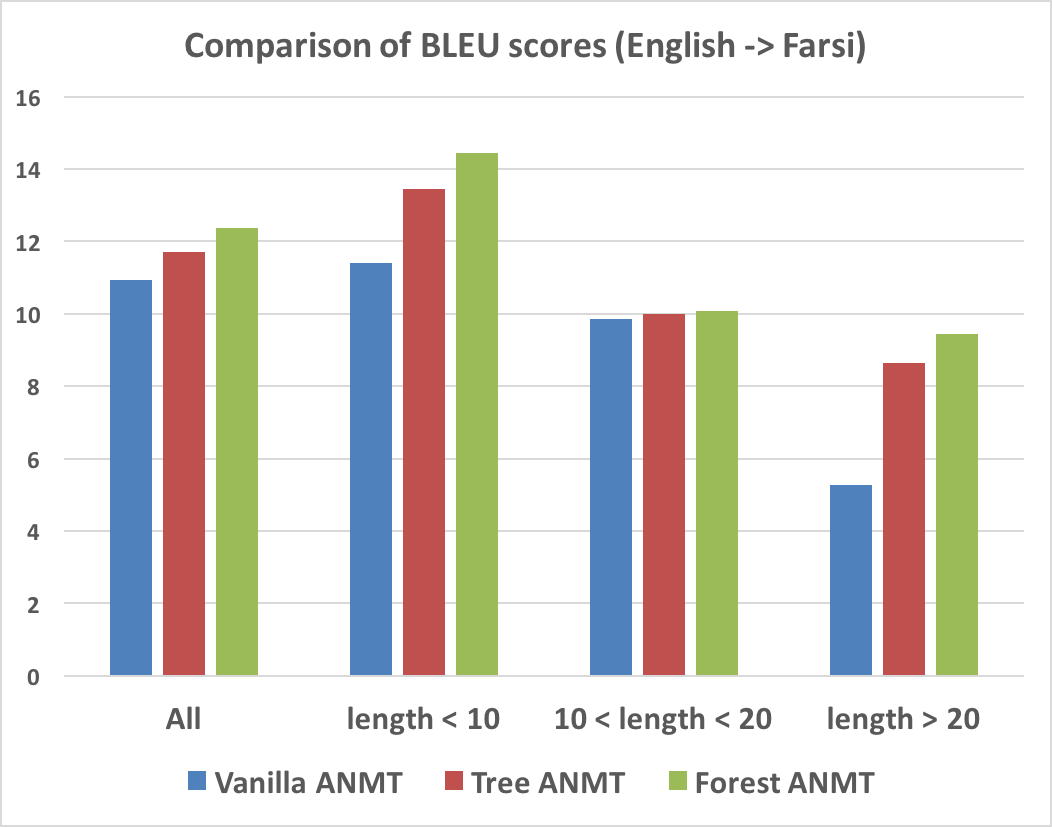}}
\caption{Comparison of BLEU scores for datasets bucketed based on the length of source sentences.}\label{fig:compare}
\end{figure*}

To analyse the effect of incorporating parse trees we investigate the effect of length of source sentences. We bucket each dataset into three parts. The first part contains pairs which source sentences have lengths equal or less than 10, the second part contains source sentences which their length lie between 10 and 20, and the last part has source sentences with more than 20 words. \Cref{fig:bf,fig:bg,fig:bc} show the BLEU score of all models on bucketed datasets. One may expects that the forest model would be just beneficial for longer sentences because the number of possible parse trees are higher. However, this results shows that the forest model can be beneficial for sentences from all lengths. On the other hand, the highest improvement belongs to the English $\rightarrow$ Farsi dataset which is the noisiest one, and has the highest uncertainty in the parsing. Therefore, it can be concluded that the forest model is the most beneficial when there is ambiguity in the sentence regardless of its size.

\begin{figure}
\centering
\includegraphics[scale=0.3]{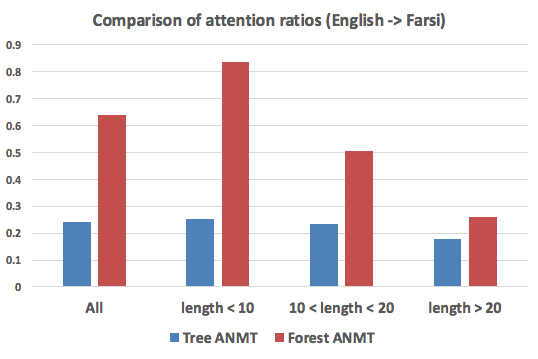}
\caption{Comparison of attention ratios for English $\rightarrow$ Persian bucketed dataset}\label{fig:forestplot}
\end{figure}

In order to analyse how models use information provided by parse trees, we compute ratio of attention on tree/forest part to attention on words part for both tree-to-sequence and forest-to-sequence models. For each sentence we calculate sum of attention on words and phrases for all target words. Then, the ratio of attention on phrases to words are computed and is averaged for all sentences. Figure \ref{fig:forestplot} shows this attention ratios for bucketed English $\rightarrow$ Persian dataset. Forests provide richer phrase embedding part since the amount of attention on forest computed embeddings is tangibly higher than tree ones.

\section{Conclusion}
We have proposed a forest-based Attentional NMT model which uses a packed forest instead of the 1-best tree in the encoder. Although using the phrase structure tree for computing embedding of phrases of source sentence enhance the accuracy of translation, parsing errors are inevitable. Using forest of parse tress and dynamic programming, our method efficiently consider exponentially many grammar trees in order to compensate for parsing errors. Experimental results showed our method is superior to the tree-to-sequence ANMT model.

\section{Acknowledgment}
This research was partly supported by CSIRO's Data61. We would like to thank Wray Buntine  and Bin Li  for fruitful discussions. 
%\bibliography{acl2017}
\bibliography{acl2017}

\begin{thebibliography}{}
\expandafter\ifx\csname natexlab\endcsname\relax\def\natexlab#1{#1}\fi

\bibitem[{Charniak and Johnson(2005)}]{charniak2005coarse}
Eugene Charniak and Mark Johnson. 2005.
\newblock Coarse-to-fine n-best parsing and {MaxEnt} discriminative reranking.
\newblock In {\em Proceedings of the 43rd annual meeting on association for
  computational linguistics\/}. Association for Computational Linguistics,
  pages 173--180.

\bibitem[{Cho et~al.(2014)Cho, Van~Merri{\"e}nboer, Gulcehre, Bahdanau,
  Bougares, Schwenk, and Bengio}]{cho2014learning}
Kyunghyun Cho, Bart Van~Merri{\"e}nboer, Caglar Gulcehre, Dzmitry Bahdanau,
  Fethi Bougares, Holger Schwenk, and Yoshua Bengio. 2014.
\newblock Learning phrase representations using {RNN} encoder-decoder for
  statistical machine translation pages 1724--1734.

\bibitem[{Cohn et~al.(2016)Cohn, Hoang, Vymolova, Yao, Dyer, and
  Haffari}]{cohn2016incorporating}
Trevor Cohn, Cong Duy~Vu Hoang, Ekaterina Vymolova, Kaisheng Yao, Chris Dyer,
  and Gholamreza Haffari. 2016.
\newblock Incorporating structural alignment biases into an attentional neural
  translation model.
\newblock {\em Proceedings of the 2016 Conference of the North American Chapter
  of the Association for Computational Linguistics\/} .

\bibitem[{Eriguchi et~al.(2016)Eriguchi, Hashimoto, and
  Tsuruoka}]{eriguchi2016tree}
Akiko Eriguchi, Kazuma Hashimoto, and Yoshimasa Tsuruoka. 2016.
\newblock Tree-to-sequence attentional neural machine translation.
\newblock In {\em Proceedings of the 54th Annual Meeting of the Association for
  Computational Linguistics\/}. Association for Computational Linguistics,
  pages 823--833.

\bibitem[{Hochreiter and Schmidhuber(1997)}]{hochreiter1997long}
Sepp Hochreiter and J{\"u}rgen Schmidhuber. 1997.
\newblock Long short-term memory.
\newblock {\em Neural computation\/} 9(8):1735--1780.

\bibitem[{Huang(2008)}]{huang2008forest}
Liang Huang. 2008.
\newblock Forest reranking: Discriminative parsing with non-local features.
\newblock In {\em ACL\/}. pages 586--594.

\bibitem[{Koehn et~al.(2007)Koehn, Hoang, Birch, Callison-Burch, Federico,
  Bertoldi, Cowan, Shen, Moran, Zens, Dyer, Bojar, Constantin, and
  Herbst}]{moses}
Philipp Koehn, Hieu Hoang, Alexandra Birch, Chris Callison-Burch, Marcello
  Federico, Nicola Bertoldi, Brooke Cowan, Wade Shen, Christine Moran, Richard
  Zens, Chris Dyer, Ond\v{r}ej Bojar, Alexandra Constantin, and Evan Herbst.
  2007.
\newblock \href{http://dl.acm.org/citation.cfm?id=1557769.1557821}{Moses: Open
  source toolkit for statistical machine translation}.
\newblock In {\em Proceedings of the 45th Annual Meeting of the Association for
  Computational Linguistics on Interactive Poster and Demonstration
  Sessions\/}. Association for Computational Linguistics, Stroudsburg, PA, USA,
  ACL '07, pages 177--180.
\newblock
  \href{http://dl.acm.org/citation.cfm?id=1557769.1557821}{http://dl.acm.org/citation.cfm?id=1557769.1557821}.

\bibitem[{Luong et~al.(2015)Luong, Pham, and Manning}]{luong2015effective}
Minh-Thang Luong, Hieu Pham, and Christopher~D Manning. 2015.
\newblock Effective approaches to attention-based neural machine translation.
\newblock In {\em Proceedings of the 2015 Conference on Empirical Methods in
  Natural Language Processing (EMNLP)\/}. Association for Computational
  Linguistics.

\bibitem[{Neubig et~al.(2017)Neubig, Dyer, Goldberg, Matthews, Ammar,
  Anastasopoulos, Ballesteros, Chiang, Clothiaux, Cohn, Duh, Faruqui, Gan,
  Garrette, Ji, Kong, Kuncoro, Kumar, Malaviya, Michel, Oda, Richardson,
  Saphra, Swayamdipta, and Yin}]{dynet}
Graham Neubig, Chris Dyer, Yoav Goldberg, Austin Matthews, Waleed Ammar,
  Antonios Anastasopoulos, Miguel Ballesteros, David Chiang, Daniel Clothiaux,
  Trevor Cohn, Kevin Duh, Manaal Faruqui, Cynthia Gan, Dan Garrette, Yangfeng
  Ji, Lingpeng Kong, Adhiguna Kuncoro, Gaurav Kumar, Chaitanya Malaviya, Paul
  Michel, Yusuke Oda, Matthew Richardson, Naomi Saphra, Swabha Swayamdipta, and
  Pengcheng Yin. 2017.
\newblock {DyNet}: The dynamic neural network toolkit.
\newblock {\em arXiv preprint arXiv:1701.03980\/} .

\bibitem[{Pascanu et~al.(2013)Pascanu, Gulcehre, Cho, and
  Bengio}]{pascanu2013construct}
Razvan Pascanu, Caglar Gulcehre, Kyunghyun Cho, and Yoshua Bengio. 2013.
\newblock How to construct deep recurrent neural networks.
\newblock {\em arXiv preprint arXiv:1312.6026\/} .

\bibitem[{Sutskever et~al.(2014)Sutskever, Vinyals, and
  Le}]{sutskever2014sequence}
Ilya Sutskever, Oriol Vinyals, and Quoc~V Le. 2014.
\newblock Sequence to sequence learning with neural networks.
\newblock In {\em Advances in {N}eural information {P}rocessing {S}ystems\/}.
  pages 3104--3112.

\bibitem[{Tai et~al.(2015)Tai, Socher, and Manning}]{tai2015improved}
Kai~Sheng Tai, Richard Socher, and Christopher~D Manning. 2015.
\newblock Improved semantic representations from tree-structured long
  short-term memory networks.
\newblock {\em arXiv preprint arXiv:1503.00075\/} .

\bibitem[{Tiedemann(2009)}]{tiedemann2009news}
J{\"o}rg Tiedemann. 2009.
\newblock News from {OPUS-A} collection of multilingual parallel corpora with
  tools and interfaces.
\newblock In {\em Recent advances in natural language processing\/}. volume~5,
  pages 237--248.

\end{thebibliography}
\bibliographystyle{acl_natbib}

\end{document}